\documentclass[a4paper, 10pt, conference]{ieeeconf}      
\usepackage{FG2017}

\FGfinalcopy 

\IEEEoverridecommandlockouts                              
\usepackage[pdftex]{graphicx} 
\graphicspath{{/}}
\usepackage{amsmath} 
\usepackage[linesnumbered]{algorithm2e}

\title{\LARGE \bf
Fusing Deep Learned and Hand-Crafted Features of Appearance, Shape, and Dynamics for Automatic Pain Estimation
}

\author{\parbox{16cm}{\centering
    {\large Joy Egede$^1$, Michel Valstar$^2$ and Brais Martinez$^2$}\\
    {\normalsize
    joy.egede@nottingham.edu.cn, michel.valstar@nottingham.ac.uk\\
    $^1$ School of Computer Science, University of Nottingham, Ningbo China\\
    $^2$ School of Computer Science, University of Nottingham, UK}}
}

\begin{document}

\ifFGfinal
\thispagestyle{empty}
\pagestyle{empty}
\else
\author{Anonymous FG 2017 submission\\-- DO NOT DISTRIBUTE --\\}
\pagestyle{plain}
\fi
\maketitle

\begin{abstract}

Automatic continuous time, continuous value assessment of a patient's pain from face video is highly sought after by the medical profession. Despite the recent advances in deep learning that attain impressive results in many domains, pain estimation risks not being able to benefit from this due to the difficulty in obtaining data sets of considerable size. In this work we propose a combination of hand-crafted and deep-learned features  that makes the most of deep learning techniques in small sample settings. Encoding shape, appearance, and dynamics, our method significantly outperforms the current state of the art, attaining a RMSE error of less than 1 point on a 16-level pain scale, whilst simultaneously scoring a 67.3\% Pearson correlation coefficient between our predicted pain level time series and the ground truth.
\end{abstract}

\section{INTRODUCTION}\label{sec:introduction}
\noindent Ever since the designation of pain as the “fifth vital sign” in medical diagnosis, pain assessment has been an issue of utmost importance in clinical practice \cite{stephVet5pain}. The current standard for clinical pain assessment in conscious adults is via self-report. A number of scales have been developed to assist with the measurement of pain using this standard e.g. the Numerical Rating Scale (NRS)\cite{williams2000simple} and the Visual Analogue Scale (VAS) \cite{Lynch2001}. Although self-report tools have been extensively researched, validated and used in clinical settings they are still limited.

These tools are not applicable to patients who do not have the ability to articulate or describe their pain; e.g. young infants, the mentally impaired and individuals whose verbal ability is inhibited by a critical medical condition or device \cite{lucey2011automatically}. Other challenges associated with self reports include differences in patient and clinician’s definition or quantification of pain, patients attempting to mask pain or report more pain than is actually experienced, or disparities in measurement properties across scales. Studies by Williams et al.  on NRS have also shown that patients tend to avoid choosing values in the upper range when the scale uses large numbers based on the impression that higher values imply unmanageable pain \cite{williams2000simple}. They also discovered that some patients re-label the scale point to something they can better relate with before assigning a score. For example, 'worst imaginable pain' becomes 'worst pain I have ever felt'.

In cases where self-report is not applicable, pain assessment is done by proxy i.e. pain intensity is estimated by an observer based on the behavioural and physiological changes in the patient \cite{Ashraf_2009}. While useful, the approach has its limitations some of which include bias due to subjectivity, contextual factors, desensitization of proxy due to prolonged exposure to pain, training/experience etc \cite{Lynch2001} \cite{McMasterDB2012}. This has led to the development of automatic pain assessment based on audio-visual recordings of expressive behaviour using state-of-the art machine learning and digital signal processing methods. This automatic analysis of expressive behaviour altered by medical condition was recently coined as "Behaviomedics" \cite{valstarbehavmedics}.

A plethora of research has been documented on  automatic pain recognition based on a variety of data sources e.g facial expression, audio and body postures. Research efforts started at distinguishing between pain and no pain\cite{Ashraf_2009}\cite{Brahnam_2007} in data samples and gradually extended to continuous pain estimation \cite{Zafar_2014}\cite{Kaltwang_2012}\cite{McMasterDB2012}; a more valuable outcome for clinical diagnosis.

Despite current achievements, there are still open challenges with automatic pain recognition. Like most recognition systems, the performance of pain recognition models is largely dependent on the quantity and quality of data used in training. Human expression data is limited in supply to begin with but pain data is particularly difficult to obtain. Where available, there is the additional complication of a sparse representation for higher pain levels. This imbalance reduces the ability of recognition models to predict high pain intensity levels. For example, this imbalance is very evident in the popularly used UNBC McMaster pain database which contains 87.21\% of 'no pain' frames and only 17.29\% of 'pain' frames.  There is a clear need for novel methods that can achieving good pain estimation results from the limited data available.


Deep learning has successfully been applied to various computer vision problems. However, deep learned nets require massive amounts of data for good performance. This has hindered it application to automatic pain recognition. Even though deep learned features can not currently work by itself for pain recognition due to the above reasons, we hypothesise that their ability to learn features has value even when small sample sizes are available for training. We will show that such learned features contain valuable information that is complementary to handcrafted features. In this work, we explore deep learning for continuous pain intensity estimation in the face of limited data. We show that combining  deep learned features with handcrafted features yields significant improvement in automatic pain recognition compared to using only the former. To the best of our knowledge our work is the first to use deep learned features for continuous pain estimation.

Secondly, good facial expression analysis uses a combination of shape based and appearance features. In this work, We encode shape and appearance information in both the hand crafted and deep learned features.  We achieve this in our deep learned features by learning features not only from the original images but also from binary image masks defined by a set of facial point locations. Thus our deep learned features are learned from a combination of original image  pixels (appearance) and the binary masks (face shape).

Furthermore, learning facial expression from static features is not sufficient. Encoding dynamic information of facial actions significantly improves the recognition performance. In this study, we encode dynamic information in the deep learned features by ensuring that features are learned from a sequence of input images defined as a specified time window centered on the current frame being analysed. 

Lastly, we adopt person-specific adaptive post processing techniques and show how they can be applied to boost the base performance of pain estimation models. We evaluate our method on the UNBC McMaster database and compare with previous studies based on handcrafted features. Our results show a significant improvement over the state of the art, and show that for a small sample size, combinations of hand-crafted and learned features obtain highest performance.

The remainder of the paper is structured as follows: section 2 presents a review of previous work in automatic pain and AU detection in general. Section 3 describes our proposed methodology of fusing deep learned features with hand crafted features continuous pain estimation. In section 4, we describe the pain database used in this work and show the experimental results in comparison to current studies. In section 5, we discuss the limitations of the PSPI metric in the light of clinical pain indicators and possible ways of incorporating other indicators to achieve a more representative pain score. We also discuss the limitations of current performance measures used in pain recognition.

\section{RELATED WORK}
\noindent Automatic pain recognition has attracted significant attention especially with its potential application to clinical diagnosis. Attempts have been made to detect/estimate pain from facial actions, head/body movement and sound analysis. Regarding pain detection from facial actions, face images have been classified into ‘pain’ or ‘no pain’ categories using a variety of classifier algorithms and face representation models. Recently, emphasis has shifted from binary pain classification to pain intensity estimation due to it's potential application clinical pain assessment. In this section, we discuss current studies on pain estimation with respect to the data sources used, pain metric employed and the machine learning techniques applied.

\subsection{PSPI based Pain Estimation}
\noindent The face is one important medium for conveying emotions. Pain as an emotion can almost entirely be judged from facial expression changes. Pain recognition based on facial actions predominantly utilises the Facial Action Coding System (FACS) developed by Ekman and Friesman\cite{hager2002facial}. FACS consists of 32 action units (AU) which are related to the movement of certain facial muscles. Specifically, 9 of these are associated with the upper face, 18 correspond to the lower face while the remaining 5 cannot be classified as either lower or upper face AUs. Pain can be described in terms of the action units activated when the pain is felt. Based on this, Prkachin and Solomon \cite{Prkachin_2008} proposed the Prkachin and Solomon pain intensity (PSPI) metric which measures pain as a linear combination of the intensities of facial action units associated with pain. (See Eqn. \ref{eq:PSPI_Eqn}). PSPI scores are assigned to images on a frame by frame bases using the metric in Eqn \ref{eq:PSPI_Eqn}. Based on this metric, a couple of studies have attempted pain recognition both at a  sequence level and at a frame level.

\begin{multline}
    \label{eq:PSPI_Eqn}
    PSPI  = AU4 + max(AU6; AU7) + \\ max(AU9; AU10)  + AU43
\end{multline}

In terms of binary pain classification, Ashraf et.al \cite{Ashraf_2009} predicted pain in patients with shoulder injuries using a combination of shape and Active Appearance Models (AAM) both at frame and sequence levels. Lucey et.al \cite{lucey2011automatically} extended this further by using non-rigid normalized 3D-AAMs to tackle the problem of spontaneous head movements associated with pain. Head movements were represented by pitch, yaw and roll computed from 3D parameters derived from the AAM. Similar to \cite{Ashraf_2009} they found that fusing the shape and appearance features yielded significant performance improvement.Taking this further, researchers have attempted to distinguish between real and posed pain \cite{Bartlett_2014}  \cite{Littlewort_2009}.

With the increased incorporation of automation in medical practice, pain recognition evolved from mere binary classification to continuous pain estimation.  Lucey, et al.\cite{McMasterDB2012}  used facial expressions and 3D head pose changes to estimate pain on a sequence level. Kaltwang, et al. \cite{Kaltwang_2012}  proposed a three-step approach for frame based pain estimation. In the first step, shape based and appearance based features are extracted from the face image. Next, separate Relevance Vector regressors are trained for each features type. The output of the RVRs are combined and used to train a second level RVR.  In contrast to AAM, feature fusion and multi-layer classifiers, Zafar and Khan \cite{Zafar_2014}   used geometric features extracted from 22 facial points and a single step K-NN classifier. A limitation of their approach is that it requires a prior annotation of the neutral face for each subject. Following the appreciable performance achieved with using non-rigid AAMs features \cite{Ashraf_2009}\cite{lucey2011automatically} to combat the problem of head movements associated with pain,  Rathe and Ganotra \cite{rathee2015novel} proposed the use of thin plate spline (TPS) to model facial action changes. TPS was used to achieve extraction of non-rigid facial features independent of their rigid or affine counterparts. A distance learning metric(DML) approach was used to map TPS deformation features to a higher discriminative space such that features from the same pain intensity level are grouped close to each other while those from other pain levels are separated as far as possible. Neshov and Manolova used supervised descent method (SDM) in combination with SIFT feature extraction for both binary and continuous pain detection. All of the aforementioned frame-based pain studies have focused on static features only. Good facial expression analysis requires a combination of static and dynamic features. Recently, Kaltwang et.al \cite{kaltwang2015doubly} combined dynamic features with part based feature extraction for continuous pain estimation. Here, the face is divided into a grid of S x S patches. Local Binary Pattern (LBP) features are extracted from each of this patches in time windowed manner. The time-scaled features from the patches are used to learn a doubly sparse RVM for pain estimation.

Deep learning has been applied to various computer vision problems but has received less attention in pain recognition. Although it has been used in \cite{chang2016application} to classify infant cries into pain, hunger and sleep, deep learning is yet to be applied to continuous pain estimation from facial expression changes. This is mostly due to the limited pain data available for learning such data intensive neural nets. In this work, we explore convolutional neural networks (CNN) for continuous pain estimation in the face of limited data. We embed dynamic information in our deep learned features by ensuring that features learned for a reference frame includes information from both preceding and subsequent frames using a defined time window. This is in contrast to Kaltwang et.al \cite{kaltwang2015doubly} where only information from preceding frames are considered.

\subsection{Non-PSPI based Pain Estimation}
\noindent Though action unit based pain recognition has witnessed significant advancements, there are still challenges that have inhibited its practical use in clinical settings. Some of these include  poor recognition rates due to out of plane head movements, illumination changes, inaccessible faces in the case of newborns in ICUs and the general aversion to being 'watched' by cameras. To mitigate this problems, a number of studies have explored  pain recognition from other pain indicators such as audible sounds or cry, body movement and physiological signals. Contextual variables has also received attention in emotion recognition\cite{Hammal2012} based on the idea that the same facial expression can have different connotations depending on the current scenario. However, not much has been done in this area due to the difficulty involved with capturing and measuring such data.

Sound or cry analysis is mostly common in newborns as it is a major form of expression at that age. Acoustic characteristics have been analysed to discriminate between normal and pain induced cry in newborns\cite{jam2009system}. Compared to pain analysis from audio-visual signals, physiological signals have received less attention. They have been used for valence and arousal detection of other emotions (e.g. joy, sadness, anger and pleasure) 
but only a few have applied it to pain detection \cite{werner2013towards} \cite{treister2012differentiating} \cite{werner2014automatic}. The use of bio-signals is still far from practical use because it is still difficult to map physiological patterns to specific emotions. Nonetheless, physiological signals are reputed to be robust to intentional emotion suppression since they are directly controlled by the nervous system and again, information can be collected in real time using bio-sensors \cite{wagner2005physiological}. Physiological signals commonly used include  galvanic skin response (GSR), photoplethysmogram (PPG), electrocardiogram (ECG), respiration changes and skin conductivity.

Attempts have also been made to combine visual features with physiological signals in  \cite{werner2014automatic}. Their findings show that using combined data sources yields better performance than individual sources. However, rather than pain intensity estimation, only a pair wise pain classification  was achieved for the four pain levels contained in the Bio-vid pain database.

\section{METHODOLOGY}
\noindent In this work we combine hand crafted features with deep learned features for pain intensity estimation. First, we extract the hand crafted features based on the 66 facial landmarks. Then we extract deep learned features from AU recognition CNNs. Next, we learn a linear regressor on the individual features and apply the RVM fusion  technique in \cite{Kaltwang_2012} for the feature combinations.

\begin{figure*}[thpb]
\centering
\includegraphics{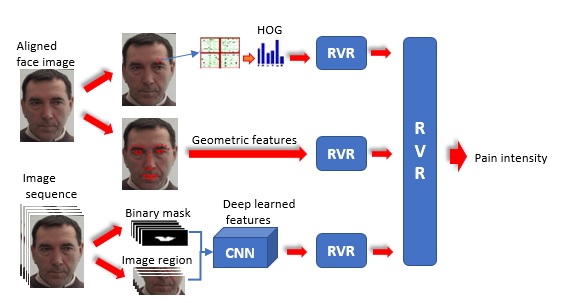}
\caption{Process flow of our proposed methodology}
\label{fig:process flow}
\end{figure*}

\subsection{Deep Learned Feature Extraction}\label{sec:deep learned features}
\noindent Considerable advancements have been recorded with deep learning in AU detection. Based on the PSPI metric, facial pain expression can be represented as combination of AUs. Consequently, it follows that features learned for AU detection can also be useful for pain estimation.
In this work, we adopt the AU detection CNN architecture proposed by Jaiswal and Valstar \cite{jaiswal2016deep}. The CNNs are pretrained for AU detection with images from the BP4D database which has a significant sample representation for the AUs relevant to pain expression. Two CNNs are trained to detect AUs associated with the eye region and mouth region of the face respectively. The CNN architecture and training process is the same as \cite{jaiswal2016deep}.

First, we extract the face from the image using the facial landmarks supplied with the database. Then the extracted face is aligned to a mean shape based on a Procrustes transform of the facial points corresponding the eye and mouth corners. These facial points are used because they are not affected by facial expression changes. We then compute a binary image mask for the face image by defining two  rectangular regions; one around the the eye and the other around the mouth using selected facial points. Next, the selected facial points are linked together in a predefined order to form a polygon. Then we generate a binary mask by setting all the points that fall within the polygons to 1 and all that fall without to 0. We denote the image region as $I$ and the corresponding binary mask as $B$ (See Fig. \ref{fig:process flow}).

To include temporal information to the learning process, the feature representation for an image at time $t$ includes information from both preceding and subsequent frames. For an image $I$ at time $t$, we retrieve a sequence of images $[I_\text{t-2},...I_t,...I_\text{t+2}]$.
Next, we extract a sequence of images $F$ such that:
\begin{equation}
    F = \left\{
    \begin{array}{rcl}
        I_j, & for & j = t \\
        I_j - I_t, & for &j\neq t.
    \end{array} \right.
\end{equation}

Similarly, for the corresponding binary mask at time $t$, we retrieve a sequence of masks $[B_\text{t-2},...,B_t,...B_\text{t+2}]$  and extract a sequence of binary masks $M$ such that;
\begin{equation}
    M = \left\{
    \begin{array}{rcl}
        B_j, & for & j = t \\
        B_j - I_t, & for &j\neq t.
    \end{array} \right.
\end{equation}

We embed dynamic information by taking the difference between the current image and (i) the next two consecutive frames (ii) the preceding two frames using a temporal window of $T=5$. The combined sequence of F and M are used as input to the CNN. The network is trained with a logarithmic loss function. 
Finally we use the pre-trained AU detection CNNs for feature extraction on the McMaster database images. The McMaster input sequences are first normalized by subtracting it from the average image of the pre-trained net. Then we retrieve the  the output of the last fully connected convolution layer (3072D) as our deep learned features. Lastly, the features from the two CNNS (i.e. eye and mouth region CNNs) are combined to give a 6144D feature set.


\subsection{Appearance and Shape based Feature Extraction}\label{sec:Appearance and Geometric features}
\noindent Appearance and shape based features have been successfully applied to automatic pain recognition particularly when used in combination. This is because they both capture unique facial characteristics which complement each other when combined. Appearance features represent subtle facial deformations caused by pain such as wrinkles and nasolabial furrow. Geometric features represent the shape and location of facial components such as the eyes, mouth, eye brow etc which are also affected by pain expression.

To extract the hand-crafted features, the face image is first pre-processed as described in section \ref{sec:deep learned features}.
For the geometric features, a number of metrics were extracted from the 49 facial points corresponding to the eyes, nose and mouth to generate a 218D feature. 

The metrics computed for each frame are as follows:
\begin{enumerate}
  \item The difference between the registered facial points and the mean shape computed from the database (98D)
  \item Euclidean distances between consecutive facial points of the eyes and eyebrow (20D)
  \item Euclidean distances between consecutive facial points of the mouth (17D)
  \item The distance between each of the facial points and the median of the stable points (49D). Stable points here refers to the landmarks corresponding to the nose and eye corners.
  \item Magnitude of the angle between three consecutive points on the eye and eye brow (18D)
  \item Magnitude of the angle between three consecutive points on the mouth (16D)
\end{enumerate}

Similarly, HOG features are extracted based on the facial point locations. We extract a patch of 24x24 pixels around each facial point. The patch is further divided into 2x2 window of cells. 
Next, 9 bins of oriented gradients are extracted from each cell thus generating a 2376D feature vector for the 66 facial points. We experimented with a number of patch sizes but the 24x24 pixel size gave the best results for our input image size.

\subsection{Pain Intensity Estimation}\label{sec:Pain Intensity Estimation}
\noindent In order to learn a regression model to estimate pain intensity, we use a Relevance Vector Regressor(RVR). RVRs have previously been used for pain intensity estimation and have been shown to perform well \cite{Kaltwang_2012}\cite{kaltwang2015doubly}. Furthermore, in comparison to SVMs which have also been widely used in facial expression analysis, RVMs use sparser models and as such train much faster \cite{bishop2006pattern}. RVMs are also less prone to over-fitting compared to SVMs \cite{xiang2007classification}. 
Indeed in our case, the number of support vectors was reduced by an average factor of 0.028 in the corresponding RVM. Table \ref{tab:model_sparsity_comparison} shows a comparison of RVM and SVM with respect to the number of selected vectors across all feature types.
\begin{table}[t]
\centering
\caption{SVM vs RVM model sparsity comparison on 16605 training samples}
\begin{tabular}{p{2.8cm} p{0.5cm} p{0.5cm} p{0.5cm} p{1cm} p{0.5cm}}
\hline
Features & GF& HOG & CNN & HOG{\_}GF & all  \\
\hline
No. of Support vectors &  8184 & 13309 & 9255 & 11882 & 11,589   \\
No. of Relevant vectors & 409 & 439 & 501 & 11 & 17   \\
\hline
\end{tabular}
\label{tab:model_sparsity_comparison}
\end{table}

First, we learn an RVR each on the geometric, HOG and CNN features. The ground truth for the RVR training are the PSPI scores corresponding to the input frame.  Previous studies\cite{Kaltwang_2012}\cite{lucey2011painful} have shown that significant performance improvement can be obtained by combining features from different sources. Hence, following \cite{Kaltwang_2012}'s technique, we combine the output of the single feature RVRs and use this to learn a second level RVR. If we denote the output of the single RVR as $r$, the input to the second level RVR $R$ is the feature set $[r_1,...r_n]$ where $n$ is the number of single feature RVRs to be combined. The bias parameter for the RVM is determined by an inner loop subject independent cross validation on the training data. RVR performance is evaluated using a leave-one-subject out cross validation. That is, at each point all the frames from one subject are used for testing while frames from all the other subjects are used for training. This is repeated until all the subjects are used for testing. To reduce the imbalance in the training set, we under-sampled the no-pain signal frames using a ratio of 1:2 for the highest occurring non-zero pain frames and the no-pain signal frames respectively. No modifications were made to the test set.

 \subsection{Post-processing of Results}\label{sec:Post-processing of Results}
\noindent After obtaining the initial predictions from the RVMs, we experimented with a number of result post processing methods to see the impact on the RVR performance. Here, we tried three different techniques. Firstly, since the RVR was capable of predicting values less than and above the minimum and max pain levels(i.e. 0 and 16), we set all predictions below the max pain level to zero and predictions above the max to 16. This technique is designated as 'thresholding'. In the second method, we computed the modal prediction for each subject and then subtracted this value from the RVM predictions for the subjects. This is denoted as 're-based'. Rebasing can be applied in a practical situation as it does not depend on any ground truth. This can be done by taking the mode over a defined time window of frames and subtracting this value from all the predictions within the time window. Lastly, we also applied thresholding on the re-based results.

\section{EXPERIMENTAL EVALUATION}
\subsection{Database description}
\begin{figure}[b]
\centering
\includegraphics[width=\linewidth]{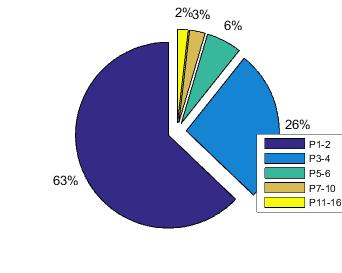}
\caption{Percentage Distribution of non-zero pain levels in the McMaster database}
\label{fig:PSPI_dist}
\end{figure}

\noindent To evaluate our proposed method, we used the publicly accessible UNBC-McMaster shoulder pain expression archive database. It consists of 200 video sequences of facial expression of 25 subjects undergoing different range of motion test i.e. abduction,  external and internal rotation of the arm. An active and passive approach was used in the data collection. In active mode, Subjects were asked to move the affected arms themselves to a bearable limit while the physiotherapist did the movements in the passive mode. Each video sequence consists of approximately 60 to 700 frames resulting to  a total of 48398 frames. 82.71\% of all the frames have a pain score of zero (0) indicating high imbalance in the positive versus negative frame contribution. Fig. \ref{fig:PSPI_dist}  shows the percentage distribution of the positive frames.

All frames in the video sequences are FACs coded for the pain related action units i.e. AU4, AU6, AU7, AU9, AU10, AU12, AU20, AU25, AU26, AU27 and AU43. Each action unit is coded on intensity of A-E or 0-5 except for AU43 (closed eyes) which has only two states: present or absent. Using the PSPI metric, a pain score is assigned based on the intensity of the AUs present.

The computed PSPI score is used as the ground truth in our RVR experiments.  In addition, the database provides 66 facial points for each frame based on an active appearance model. The facial points provided are used in this study for HOG and geometric feature extraction.

\subsection{Experiments and Results}
\noindent To support comparison with previous work, we used the Pearson correlation (CORR) and root mean square error (RMSE) to evaluate performance. Even though MSE is used in previous studies, we use the RMSE because it is on the same scale as the predictions and easier to interpret. Performance is computed by first concatenating the predictions on all the subjects frames. RMSE is computed as the difference between the RVM predictions and the ground truth while CORR is the Pearson correlation between the concatenated predictions and the ground truth.

We also compared the effect of the prediction processing methods described in \ref{sec:Post-processing of Results} in relation to the unprocessed predictions. Tables \ref{tab:post-process_comparison} shows a comparison of the the effect of the result post processing techniques on the RVR performance for both the individual and combined features.



\begin{table*}[t]
\centering
\caption{Comparison of result post processing methods}
\bgroup
\def\arraystretch{1.1}
\begin{tabular}{l|l l l l l |l l l l l l l}
\hline
Measure & & &RMSE & & & & &CORR\\
\hline
Features & GF& HOG & CNN & HOG{\_}GF & all & GF& HOG & CNN & HOG{\_}GF & all\\
\hline
Original &1.3679 &1.2412 &1.3733& 1.3700&1.3003 &0.4729 &0.5130 &0.508	 &0.5961&0.6323\\
Rebased  &1.2807 &1.1344 &1.2541& 1.0821&1.0261 &0.4915	&0.5621	&0.4907	 &0.6300&0.6542\\
Threshholding &1.2048	&1.1278	&1.1713& 1.0596&1.1141 &0.5092 &0.5479 &0.5184 & 0.5860&0.6184\\
Rebased{\_}threshholding &\textbf{1.147}&\textbf{1.0667}& \textbf{1.1605}&\textbf{1.0248} &\textbf{0.9926} & \textbf{0.5475} &\textbf{0.609} &\textbf{0.5356} & \textbf{0.6479} &\textbf{0.6728}\\
\hline
\end{tabular}
\egroup
\label{tab:post-process_comparison}
\end{table*}

From Table \ref{tab:post-process_comparison}, it can be seen that processing the predictions results in significant performance improvement compared to the unprocessed predictions.
'Rebasing' impacts more on the correlation performance compared to the 'Thresholding' method in most cases. A possible explanation for this is that the neutral face (i.e. no pain face) constitutes a greater proportion of the videos. Thus, in cases where a non zero positive value is mostly predicted for the neutral face, subtracting this value from the frame predictions fine-tunes the neutral-face predictions towards the ground truth. This has a similar impact on the 'pain face' predictions, as subtracting the modal prediction (neutral face) leaves us with only the effect of the pain expression.
On the other hand, 'Thresholding' achieves lower RMSE scores compared to 'Rebasing'. This is because threshholding maps extreme predictions to the appropriate pain scale limits which further reduces the magnitude of the prediction error.  Combining this two methods will yield better improvements for both performance measures. As expected, the 'rebased{\_}thresholding' methods consistently outperforms all the others with a significant margin..



\begin{figure}[b]
\includegraphics[width=\linewidth, keepaspectratio]{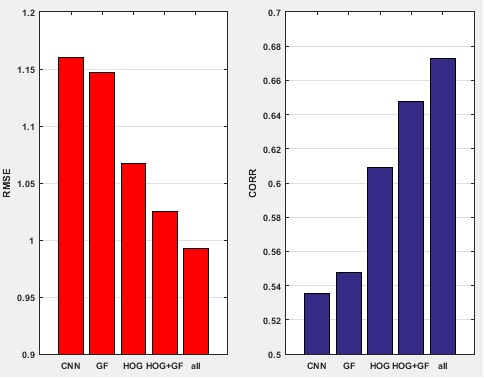}
\caption{Comparison of RVM performance on the different features}
\label{fig:feature comparison}
\end{figure}

Using the best result processing method, we compare the RVM performance on the different features in Fig.\ref{fig:feature comparison}. Among the single features, HOG performs better than all the others followed by the geometric features. The CNN features have the lowest performance. This could be explained by the fact that the CNN features have been fine-tuned to suit the original problem i.e. AU detection whereas the appearance and shaped based features are more generic in nature. This also shows that deep learned features do not work well with small data. Nonetheless, the CNN feature measures considerably well with the geometric features especially in CORR.
Similar to previous observations \cite{Kaltwang_2012} \cite{lucey2011automatically}, the appearance based features perform significantly better than the shape based features. A possible reason for this is that appearance features effectively capture facial deformation caused by pain intensity. On the other hand, facial pain expression is often accompanied by out-of plane head movement which negatively impacts on  shape registration using Procrustes alignment.

It can also be seen that the combined features  perform much better than the single features with HOG+GF+CNN (designated as 'all') performing better than hand crafted feature combination. This shows that though CNN features do not work well on their own but they provide valuable information which complement the handcrafted features. Furthermore, the CNN features are learned directly from the image pixels with less loss of information whereas the handcrafted features are learned on high level representations of the original face image and are prone to oversimplification. This added advantage could have contributed to the improved performance when all features are combined.     

Finally, in Table \ref{tab:state_of_the_art_comparison} we compared our proposed method with the state of the art. Our method outperforms the others in both MSE and CORR. Specifically, in comparison to Neshov and Manolava \cite{neshov2015pain} we achieved a 14\% increase in both performance measures. In comparison to \cite{kaltwang2015doubly} we achieved a 2\% increase in CORR and a massive 70\% reduction in RMSE. All the other methods apart from \cite{kaltwang2015doubly} used only static features whereas our method incorporates dynamic information to the learning process. This is evident in the comparable CORR obtained in \cite{kaltwang2015doubly} even though this is not the case with RMSE measure.

\begin{table}[t]
\centering
\caption{Comparison of our method with the state-of-the-art}
\begin{tabular}{p{3cm}l p{1.5cm}l l }
\hline
&RMSE&CORR  \\
\hline
Kaltwang et.al \cite{Kaltwang_2012} & 1.18 & 0.59   \\
Neshov and Manolava\cite{neshov2015pain} & 1.13 & 0.59   \\
Kaltwang et.al \cite{kaltwang2015doubly}& 1.69& 0.66\\
Our method           & \textbf{0.99} & \textbf{0.67}  \\
\hline
\end{tabular}
\label{tab:state_of_the_art_comparison}
\end{table}

\section{LIMITATION OF CURRENT PRACTICES}
\noindent A group of publications has recently addressed the problem of automatic pain estimation from face videos. Here we discuss some of the common problems that we feel need to be addressed as a community in order to make meaningful progress.

\subsection {Prkachin's pain metric}
\noindent The Prkachin's pain metric has been widely used for automatic pain recognition studies. While it is useful as an evaluation tool, it ignores some important behavioural indicators of pain. For example, it does not consider head, body movements and pain related sounds which have been identified as important indicators of pain especially in newborns\cite{Ranger2007283} \cite{lawrence1993development} \cite{lucey2011automatically}. Automatic pain recognition requires a more robust pain metric that captures all of the audio-visual expressions of pain in order for it to be applicable to clinical settings. Considering that face based automatic recognition systems are still adversely affected by out of plane head movements, the actual pain suffered by a patient can be under-judged using only a face-based metric. A multi-indicator pain metric will be more appropriate for cases where the face is occluded. Contextual pain indicators can also be captured in the metric by adding more weights to  certain indicators depending on the context of use. For example since newborns are more prone to express their pain via crying, the weight for audio indicators will be increased for such scenarios. A similar idea of using weights to capture contextual factors has been demonstrated in \cite{Hammal2012} but in this case it was the classifier that was biased to the context not the pain metric. A metric that meets the above specifications will be of more use to clinical pain assessment.

\subsection{Evaluation Metrics}
\noindent In studies on continuous pain estimation from facial features, performance evaluation has mostly been based on the (root) mean square error ((R)MSE) and the Pearson correlation coefficient (CORR). The MSE error effectively captures the difference between the model predictions and the ground truth, whereas the CORR measures how well the prediction follows the ground truth trend, irrespective of a potential absolute bias.

Both measures clearly have their own merit, but it is clear that a low RMSE isn't particularly valuable without a high CORR, and vice-versa. For example, due to the class imbalance it is fairly trivial to attain a very low RMSE by predicting all frames to have a pain level of 0.  A limitation of the CORR measure is also evident in the performance in \cite{kaltwang2015doubly} who report a very high MSE compared to other previous studies but attain a high correlation. Based on this limitation, it is necessary to find a better performance measure that is robust to data imbalance.

\begin{figure}[t]
\includegraphics[width=\linewidth, keepaspectratio]{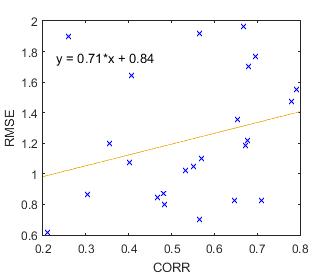}
\caption{Limitation of PCOR as a performance measure (Subject based RMSE vs CORR on HOG features.)}
\label{fig:PCOR_limitation}
\end{figure}

\begin{figure}[th!]
\includegraphics[width=\linewidth, keepaspectratio]{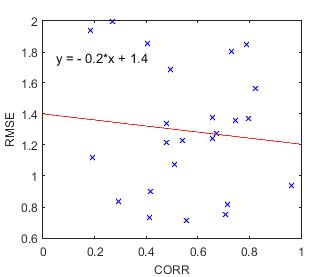}
\caption{Subject based RMSE vs CORR on HOG+GF+CNN features showing a more logical gradient.}
\label{fig:rmse_vs_pcor_fusion}
\end{figure}

More generally, we have found that there isn't necessarily a sensible correlation between good CORR and good RMSE.  Fig. \ref{fig:PCOR_limitation} shows a plot of RMSE vs CORR for the RVR prediction on HOG features. It can be observed that it is possible to attain very high CORR yet still attain high RMSE even though a high CORR should logically imply a low RMSE.  This is possible in cases where the RVM is unable to predict high pain levels well. For example, if the RVM predicts half the magnitude of the high pain levels, we get a nice correlation but a high error margin. This is particularly possible with the McMaster database which suffers from sparse data representation for high pain values. Again, we observe a positive gradient on the line of best fit whereas an ideal plot of RMSE against CORR should have a negative gradient.

Interestingly, when the same graph is plotted for the RVR prediction on the combination of handcrafted and deep learned features (see Fig. \ref{fig:rmse_vs_pcor_fusion}), we observe a more logical negative gradient for the line of best fit. This implies that the feature combination somewhat combats the problem of data imbalance as the graph now tends towards the expected gradient. Possibly this means that a tipping point has been reached in terms of performance, but this should be confirmed empirically.

\section{CONCLUSIONS AND FUTURE WORKS}
\noindent We have introduced a method that combines deep learned features with hand crafted features for continuous pain estimation. We encode shape and appearance information in our deep learned features by generating a binary image mask based on the facial landmarks. Dynamic information is embedded to the deep learned features using a defined time window. We show that our proposed method of combining handcrafted features with deep learned features  yields significant improvement over the former and outperforms the state of the art. Our system being a face based approach is still limited in the sense that it requires near frontal faces to work well. In our future work, we will look at pain estimation from a combination of pain indicators in addition to facial expressions.


\bibliography{mybib}

\begin{thebibliography}{10}

\bibitem{Ashraf_2009}
Ahmed~Bilal Ashraf, Simon Lucey, Jeffrey~F. Cohn, Tsuhan Chen, Zara Ambadar,
  Kenneth~M. Prkachin, and Patricia~E. Solomon.
\newblock {The painful face {\textendash} Pain expression recognition using
  active appearance models}.
\newblock {\em Image and Vision Computing}, 27(12):1788--1796, nov 2009.

\bibitem{Bartlett_2014}
Marian~Stewart Bartlett, Gwen~C. Littlewort, Mark~G. Frank, and Kang Lee.
\newblock {Automatic Decoding of Facial Movements Reveals Deceptive Pain
  Expressions}.
\newblock {\em Current Biology}, 24(7):738--743, mar 2014.

\bibitem{bishop2006pattern}
Christopher~M Bishop et~al.
\newblock Pattern recognition and machine learning, vol. 1.
\newblock (4):345, 2006.

\bibitem{Brahnam_2007}
Sheryl Brahnam, Chao-Fa Chuang, Randall~S. Sexton, and Frank~Y. Shih.
\newblock {Machine assessment of neonatal facial expressions of acute pain}.
\newblock {\em Decision Support Systems}, 43(4):1242--1254, aug 2007.

\bibitem{chang2016application}
Chuan-Yu Chang and Jia-Jing Li.
\newblock Application of deep learning for recognizing infant cries.
\newblock In {\em 2016 IEEE International Conference on Consumer
  Electronics-Taiwan (ICCE-TW)}, pages 1--2. IEEE, 2016.

\bibitem{hager2002facial}
Joseph~C Hager, Paul Ekman, and Wallace~V Friesen.
\newblock Facial action coding system.
\newblock {\em Salt Lake City, UT: A Human Face}, 2002.

\bibitem{Hammal2012}
Zakia Hammal and Miriam Kunz.
\newblock {Pain monitoring: A dynamic and context-sensitive system}.
\newblock {\em Pattern Recognition}, 45(4):1265--1280, 2012.

\bibitem{jaiswal2016deep}
Shashank Jaiswal and Michel Valstar.
\newblock Deep learning the dynamic appearance and shape of facial action
  units.
\newblock In {\em 2016 IEEE Winter Conference on Applications of Computer
  Vision (WACV)}, pages 1--8. IEEE, 2016.

\bibitem{jam2009system}
Mahmoud~Mansouri Jam and Hamed Sadjedi.
\newblock A system for detecting of infants with pain from normal infants based
  on multi-band spectral entropy by infant's cry analysis.
\newblock In {\em Computer and Electrical Engineering, 2009. ICCEE'09. Second
  International Conference on}, volume~2, pages 72--76. IEEE, 2009.

\bibitem{Kaltwang_2012}
Sebastian Kaltwang, Ognjen Rudovic, and Maja Pantic.
\newblock {Continuous Pain Intensity Estimation from Facial Expressions}.
\newblock In {\em Advances in Visual Computing}, pages 368--377. Springer
  Science + Business Media, 2012.

\bibitem{kaltwang2015doubly}
Sebastian Kaltwang, Sinisa Todorovic, and Maja Pantic.
\newblock Doubly sparse relevance vector machine for continuous facial behavior
  estimation.
\newblock 2015.

\bibitem{lawrence1993development}
Jocelyn Lawrence, Denise Alcock, Patrick McGrath, J~Kay, S~Brock MacMurray, and
  C~Dulberg.
\newblock The development of a tool to assess neonatal pain.
\newblock {\em Neonatal network: NN}, 12(6):59--66, 1993.

\bibitem{Littlewort_2009}
Gwen~C. Littlewort, Marian~Stewart Bartlett, and Kang Lee.
\newblock {Automatic coding of facial expressions displayed during posed and
  genuine pain}.
\newblock {\em Image and Vision Computing}, 27(12):1797--1803, nov 2009.

\bibitem{lucey2011automatically}
Patrick Lucey, Jeffrey~F Cohn, Iain Matthews, Simon Lucey, Sridha Sridharan,
  Jessica Howlett, and Kenneth~M Prkachin.
\newblock Automatically detecting pain in video through facial action units.
\newblock {\em Systems, Man, and Cybernetics, Part B: Cybernetics, IEEE
  Transactions on}, 41(3):664--674, 2011.

\bibitem{McMasterDB2012}
Patrick Lucey, Jeffrey~F Cohn, Kenneth~M Prkachin, Patricia~E Solomon, Sien
  Chew, and Iain Matthews.
\newblock {Painful monitoring: Automatic pain monitoring using the
  UNBC-McMaster shoulder pain expression archive database}.
\newblock {\em Image and Vision Computing}, 30(3):197--205, 2012.

\bibitem{lucey2011painful}
Patrick Lucey, Jeffrey~F Cohn, Kenneth~M Prkachin, Patricia~E Solomon, and Iain
  Matthews.
\newblock Painful data: The unbc-mcmaster shoulder pain expression archive
  database.
\newblock In {\em Automatic Face \& Gesture Recognition and Workshops (FG
  2011), 2011 IEEE International Conference on}, pages 57--64. IEEE, 2011.

\bibitem{Lynch2001}
M~Lynch.
\newblock {Pain as the fifth vital sign}.
\newblock {\em Journal of intravenous nursing : the official publication of the
  Intravenous Nurses Society}, 24(2):85—94, 2001.

\bibitem{neshov2015pain}
Nikolay Neshov and Agata Manolova.
\newblock Pain detection from facial characteristics using supervised descent
  method.
\newblock In {\em Intelligent Data Acquisition and Advanced Computing Systems:
  Technology and Applications (IDAACS), 2015 IEEE 8th International Conference
  on}, volume~1, pages 251--256. IEEE, 2015.

\bibitem{Prkachin_2008}
Kenneth~M. Prkachin and Patricia~E. Solomon.
\newblock {The structure reliability and validity of pain expression: Evidence
  from patients with shoulder pain}.
\newblock {\em Pain}, 139(2):267--274, oct 2008.

\bibitem{Ranger2007283}
Manon Ranger, C.~Céleste Johnston, and K.J.S. Anand.
\newblock Current controversies regarding pain assessment in neonates.
\newblock {\em Seminars in Perinatology}, 31(5):283 -- 288, 2007.
\newblock Pain.

\bibitem{rathee2015novel}
Neeru Rathee and Dinesh Ganotra.
\newblock A novel approach for pain intensity detection based on facial feature
  deformations.
\newblock {\em Journal of Visual Communication and Image Representation},
  33:247--254, 2015.

\bibitem{stephVet5pain}
Joan Stephenson.
\newblock {Veterans' pain a vital sign}.
\newblock {\em JAMA}, 281(11):978--978, 1999.

\bibitem{treister2012differentiating}
Roi Treister, Mark Kliger, Galit Zuckerman, Itay~Goor Aryeh, and Elon
  Eisenberg.
\newblock Differentiating between heat pain intensities: the combined effect of
  multiple autonomic parameters.
\newblock {\em PAIN{\textregistered}}, 153(9):1807--1814, 2012.

\bibitem{valstarbehavmedics}
Michel Valstar.
\newblock {Automatic Behaviour Understanding in Medicine}.
\newblock In {\em Proc. Int'l Conference Multimodal Interaction}, 2014.

\bibitem{wagner2005physiological}
Johannes Wagner, Jonghwa Kim, and Elisabeth Andr{\'e}.
\newblock From physiological signals to emotions: Implementing and comparing
  selected methods for feature extraction and classification.
\newblock In {\em Multimedia and Expo, 2005. ICME 2005. IEEE International
  Conference on}, pages 940--943. IEEE, 2005.

\bibitem{werner2013towards}
Philipp Werner, Ayoub Al-Hamadi, Robert Niese, Steffen Walter, Sascha Gruss,
  and Harald~C Traue.
\newblock Towards pain monitoring: Facial expression, head pose, a new
  database, an automatic system and remaining challenges.
\newblock In {\em Proceedings of the British Machine Vision Conference}, 2013.

\bibitem{werner2014automatic}
Philipp Werner, Ayoub Al-Hamadi, Robert Niese, Steffen Walter, Sascha Gruss,
  and Harald~C Traue.
\newblock Automatic pain recognition from video and biomedical signals.
\newblock In {\em Pattern Recognition (ICPR), 2014 22nd International
  Conference on}, pages 4582--4587. IEEE, 2014.

\bibitem{williams2000simple}
Amanda C de~C Williams, Huw Talfryn~Oakley Davies, and Yasmin Chadury.
\newblock Simple pain rating scales hide complex idiosyncratic meanings.
\newblock {\em Pain}, 85(3):457--463, 2000.

\bibitem{xiang2007classification}
Xu~Xiang-min, Mao Yun-feng, Xiong Jia-ni, and Zhou Feng-le.
\newblock Classification performance comparison between rvm and svm.
\newblock In {\em 2007 International Workshop on Anti-Counterfeiting, Security
  and Identification (ASID)}, pages 208--211. IEEE, 2007.

\bibitem{Zafar_2014}
Zuhair Zafar and Nadeem~Ahmad Khan.
\newblock {Pain Intensity Evaluation through Facial Action Units}.
\newblock In {\em 2014 22nd International Conference on Pattern Recognition}.
  Institute of Electrical {\&} Electronics Engineers ({IEEE}), aug 2014.

\end{thebibliography}
\bibliographystyle{plain}





\end{document}